\pdfoutput=1
\documentclass[11pt]{article}

\usepackage{ACL2023}

\usepackage{times} 
\usepackage{latexsym}
\usepackage{multirow}
\usepackage[T1]{fontenc}
\usepackage[utf8]{inputenc}

\usepackage{microtype}

\usepackage{inconsolata}

\usepackage{mathtools}

\usepackage{enumitem}

\usepackage{xcolor}

\usepackage{url}

\usepackage{float}

\DeclarePairedDelimiter\floor{\lfloor}{\rfloor}

\title{Revisiting Zero-Shot Abstractive Summarization in the Era of Large Language Models from the Perspective of Position Bias}

\author{
Anshuman Chhabra, Hadi Askari, Prasant Mohapatra\\
Department of Computer Science, University of California, Davis\\%~\\
\texttt{\{chhabra,haskari,pmohapatra\}@ucdavis.edu}%~\\
}

\begin{document}

\maketitle
\begin{abstract}
We characterize and study zero-shot abstractive summarization in Large Language Models (LLMs) by measuring \textit{position bias}, which we propose as a general formulation of the more restrictive \textit{lead bias} phenomenon studied previously in the literature. Position bias captures the tendency of a model unfairly prioritizing information from certain parts of the input text over others, leading to undesirable behavior. Through numerous experiments on four diverse real-world datasets, we study position bias in multiple LLM models such as \textit{GPT 3.5-Turbo}, \textit{Llama-2}, and \textit{Dolly-v2}, as well as state-of-the-art pretrained encoder-decoder abstractive summarization models such as \textit{Pegasus} and \textit{BART}. Our findings lead to novel insights and discussion on performance and position bias of models for zero-shot summarization tasks. 
\end{abstract}

\section{Introduction}
\looseness-1Deep learning based abstractive text summarization models and Large Language Models (LLMs) have shown remarkable progress in generating concise and coherent summaries from input articles that are comparable to human-written summaries \citep{zhang2023benchmarking}. Building upon this research, we aim to quantitatively measure summarization performance of LLMs (and pretrained encoder-decoder models for reference) by proposing \textit{position bias}, which is a novel and general formulation of the \textit{lead bias} phenomenon \citep{liu2019text}.

Position bias refers to the tendency of models to prioritize information from certain parts of the source text, potentially overlooking crucial details in other parts of the input article. While position bias has been studied previously in the literature as \textit{lead bias}, we posit that lead bias is a specific case of position bias. Most prior works in this domain aim to propose methods that either \textit{incorporate} or \textit{alleviate} lead bias in models for improved performance \citep{xing2021demoting, zhu2021leveraging} without a thorough analysis of the problem itself. It is also important to note that a formal definition for lead bias is still currently lacking in related work. 

\begin{figure}[H]
  \centering
  \includegraphics[width=0.48\textwidth]{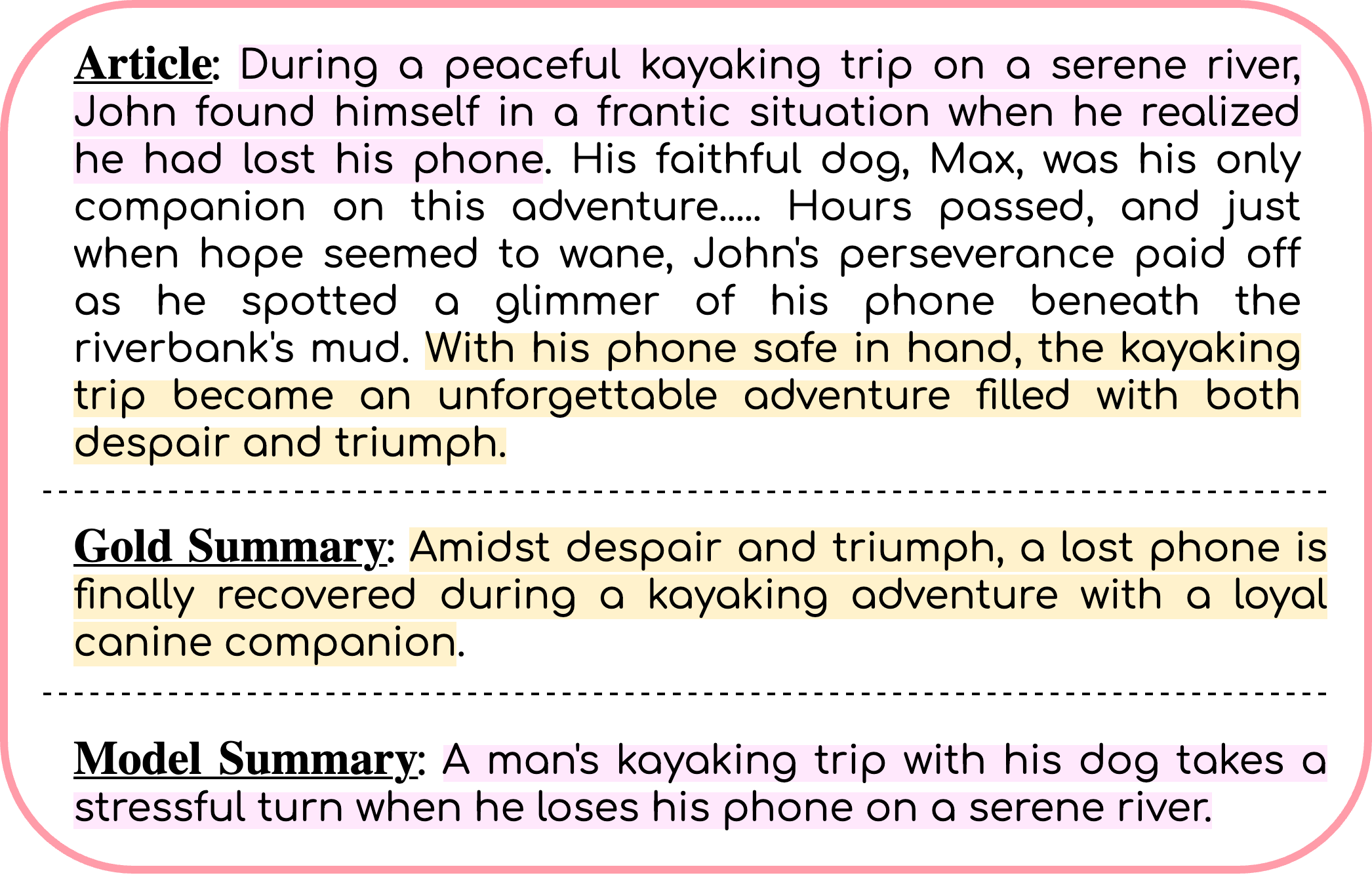}\vspace{-3mm}
  \caption{An example of \textit{position bias} where gold summary is \textit{tail biased} and model summary is \textit{lead biased}.}
  \label{fig:blurb}
\end{figure}

In contrast to lead bias, position bias seeks to decipher if models are over-utilizing sentences from any section(s) of the input articles, instead of just the leading segment. Moreover, a model's output summary can only be considered positionally biased if it overwhelmingly uses sentences from section(s) of the input that the human-written (or \textit{gold}) summaries do not use themselves. For instance, if gold summaries for a dataset are lead biased and the model generates lead biased summaries, this is desirable behavior and cannot constitute position bias. In this scenario, if the model were to generate \textit{tail biased} summaries, it would be regarded as position bias. An example of a \textit{positionally biased} summary is shown in Figure \ref{fig:blurb}.

\vspace{-1.5mm}
\vspace{-3.5mm}

We show how position bias can be empirically estimated by generating a distributional mapping between summary sentences and the article sentences used to generate the summary. Then, position bias can be measured using a metric such as Wasserstein distance \citep{vaserstein1969markov} between the model generated summary distribution and the gold summary distribution. Position bias measurements augmented with traditional metrics such as ROUGE scores \citep{lin2004rouge} can provide a more holistic evaluation of zero-shot summarization models. 

In summary, we make the following contributions in this work:

\begin{itemize}[wide,nosep]
   \item We generalize and formalize the notion of lead bias as \textit{position bias} in zero-shot abstractive summarization. Lead bias can then be understood as a specific case of position bias (Section \ref{sec:pos_bias_defn}).
   \item We show how position bias can be empirically estimated for a given zero-shot summarization model and hence, can be employed as a metric for summarization quality alongside traditional metrics such as ROUGE scores. We conduct extensive experiments to benchmark LLMs (\textit{GPT 3.5-Turbo}\footnote{\scriptsize\url{https://platform.openai.com/docs/models/gpt-3-5}}, \textit{Llama2-13B-chat}\footnote{\scriptsize\url{https://huggingface.co/meta-llama/Llama-2-13b-chat-hf}}, and \textit{Dolly-v2-7B} \citep{dolly}) and pretrained encoder-decoder models (\textit{Pegasus} \citep{zhang2020pegasus} and \textit{BART} \citep{lewis2020bart}) on 4 diverse datasets: \textit{CNN/DM} \citep{cnn}, \textit{Reddit TL;DR} \citep{reddit}, \textit{News Summary} \citep{news}, and \textit{XSum} \citep{xsum} (Section \ref{sec:results}). %Our results illustrate which models are less position biased and more performant at zero-shot summarization.
   %\item Our results suggest that on average, LLMs are less position biased and better summarizers in most cases, with GPT 3.5-Turbo being the best performer across both metrics. However, LLMs tend to exhibit more position bias in the case of \textit{extreme summarization}. Finally, we compile additional novel insights based on our findings to aid practitioners in selecting the right model for their zero-shot summarization tasks. (Section \ref{sec:discussion}).
   \item Using our findings, we compile novel insights to aid practitioners in selecting the right model for their zero-shot summarization tasks. (Section \ref{sec:discussion})
\end{itemize}

\section{Related Works}

\looseness-1Related work has studied the more specific phenomenon of lead bias in summarization. Both \citet{grenander2019countering} and \citet{xing2021demoting} propose approaches and architectural changes to models that can reduce lead bias in \textit{extractive summarization}, where summary sentences are selected directly from the source text. In contrast, in our work we study position bias more generally in \textit{abstractive summarization}. Interestingly, \citet{zhu2021leveraging} seek to leverage lead biased pre-training to improve performance on news articles, which are known to be lead biased. Prior work has also analyzed LLMs for their performance as zero-shot abstractive summarizers \citep{retkowski2023current}. \citet{goyal2022news} study GPT-3 specifically in the context of news summarization and \citet{zhang2023benchmarking} benchmark the summarization performance of multiple LLMs on the \textit{CNN/DM} and \textit{XSum} datasets. \citet{tam2023evaluating} study the factuality of summaries generated by LLMs and \citet{shen2023large} use GPT 3.5-Turbo for evaluating summaries generated by other models. In vision, model bias has been investigated for video summarization \citep{tmlr2023}. Unlike our work, none of these have studied position bias of zero-shot summarization in LLMs.

\section{Proposed Approach}

\subsection{Zero-Shot Abstractive Summarization}

A zero-shot abstractive text summarization model $\mathcal{A}$ operates on the dataset tuple $D = (X, G)$ where $X$ is a set of articles and $G$ are their corresponding reference \textit{gold} summaries, generally written by human annotators. Moreover, each article and its corresponding gold summary consists of a variable number of sentences. The model $\mathcal{A}$ then takes in as input the set of articles in the set $X$ and outputs a summary, i.e., $\mathcal{A}(X) = S$ where $S$ is the generated summary. Traditionally, the model is evaluated by comparing the generated summaries ($S$) with the gold summaries ($G$) using the ROUGE metric \citep{lin2004rouge}. We use $R^1/R^2/R^L$ to denote average ROUGE-1, ROUGE-2, and ROUGE-L scores. 

\subsection{Formulating and Estimating Position Bias}\label{sec:pos_bias_defn}
Let an article $x \in X$ have $|x| = N_x$ number of sentences. We also obtain the set of generated summaries as $S = \mathcal{A}(X)$ where each $s \in S$ has $N_s$ number of sentences. Since we consider abstractive summarization\footnote{In \textit{extractive summarization}, there is an exact one-to-one mapping between summary and article sentences.}, let us also assume we have a mapping function $\phi$ that takes in a summary sentence $s_i \in s$ and maps it back to a sentence $x_j \in x$ in the article that it was primarily derived from. Any similarity function can be employed as a useful approximation for such a mapping function $\phi$.\footnote{ROUGE or TF-IDF vector similarities are some examples.}

\begin{figure*}[!ht]
  \centering
  \includegraphics[width=0.92\textwidth]{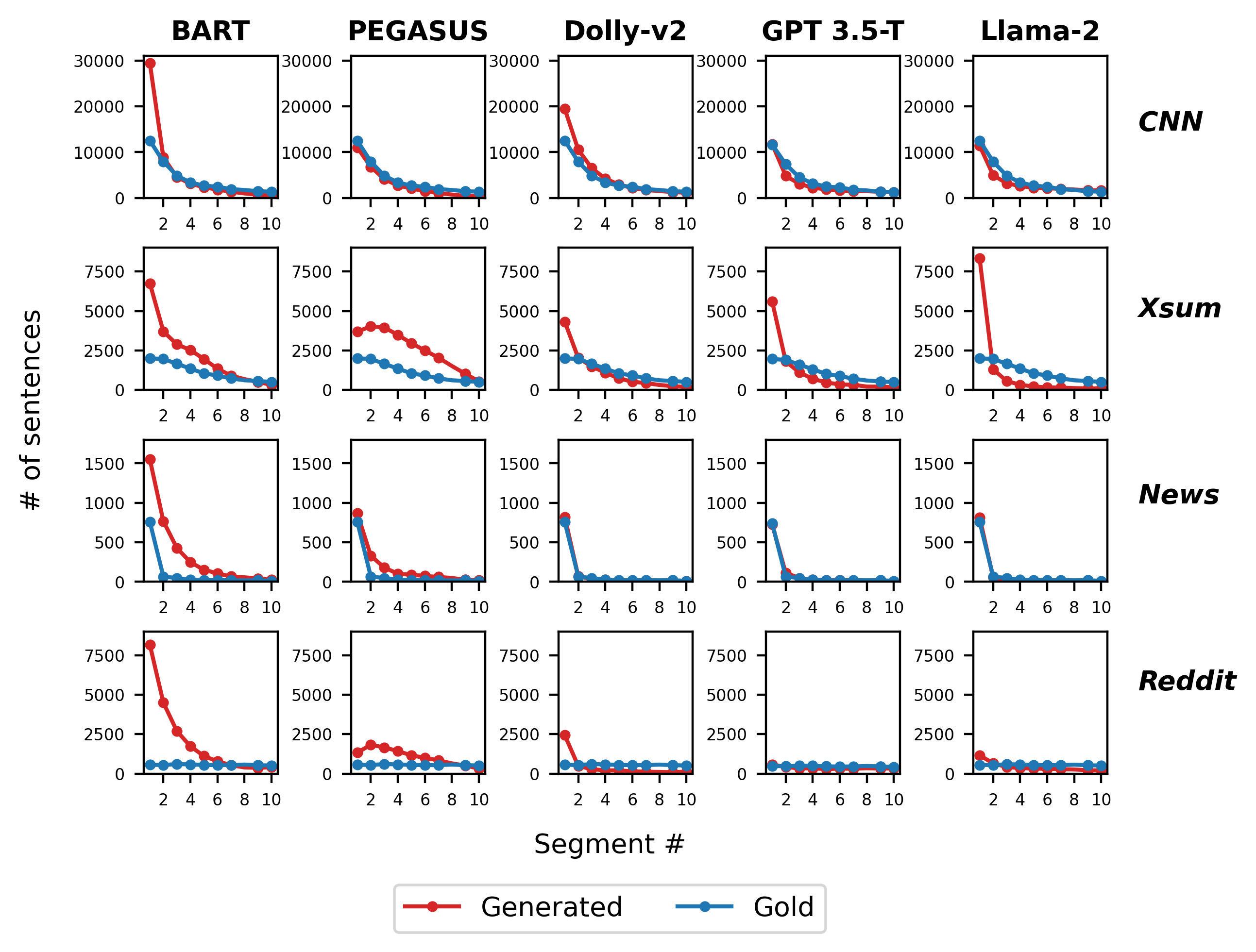}\vspace{-4mm} %0.65
  \caption{Visualizing positional distributions of gold and model generated summaries for all datasets. The more "different" these distributions are for a given dataset/model, the more \textit{position biased} the model is for that dataset.}
  \label{fig:lines}
\end{figure*}

\looseness-1Most works on lead bias implicitly assert that lead bias exists if for most $s_i \in s$, $\phi(s_i)$ maps to some $x_j$ that lies between the first $(0,k']$ sentence positions of the article. Here $k' \ll N_x$ and can be a dataset specific parameter-- for example, for the Lead-3 \citep{liu2019text} evaluation metric, $k'=3$. However, this does not seem to be a reasonable definition, especially when considering general position. For example, consider a model which tends to derive information for generating summaries by using only the last few sentences of the article. This \textit{tail bias} might also constitute undesirable behavior if the gold summaries are not tail biased themselves, but will not be accounted for in the lead bias paradigm. Hence, it is better to reason about position more generally.

%Since articles can be of differing lengths, we need better mathematical structure for describing position. Let an article $x$ be divided into $K$ segments of approximately equal length. To achieve this, the $j$-th segment will contain the sentences of the article that lie in the interval $[(j-1)\cdot c + \min(j-1,d), j\cdot c + \min(j,d)-1]$ where $K \leq \min_{x \in X}|x|$, $c = \floor*{\frac{N_x}{K}}$, and $d = N_x\mod K$. This implies that each article will have at least one sentence in a segment, and we now have a way of measuring sentence position in an article irrespective of article length.

Since articles can be of differing lengths, position becomes specific to an individual article. To overcome this issue, we divide each article $x$ into $K$ segments of approximately equal length (refer to Appendix \ref{appendix:math} for how to do this) where $K \leq \min_{x \in X}|x|$. This results in each article having at least one sentence in a segment, and we now have a uniform way of measuring sentence position across articles irrespective of their length.

\looseness-1To quantify position bias, we first obtain which article sentences the summary sentences are derived from using $\phi$, for both gold $G$ and generated $\mathcal{A}(X)$ summaries. Then, we can map these article sentences to article segments to obtain a general sense of position in the article. We now have a distributional mapping of summary sentences to article segments. Using Wasserstein distance \citep{vaserstein1969markov} between the $G$ and $\mathcal{A}(X)$ \textit{positional distributions} we can then measure position bias.

\section{Results}\label{sec:results}

For experiments, we consider the \textit{CNN/DM}, \textit{XSum}, \textit{News}, and \textit{Reddit} datasets. All datasets are different, in terms of domain, inherent position biases, or article and gold summary length. We choose only instruction-tuned LLMs as they are more performant at summarization \citep{retkowski2023current} and we cover different model sizes: GPT 3.5-T is large (175B params), Llama-2 is mid-size (13B params), and Dolly-v2 is small (7B params). We also consider SOTA pretrained encoder-decoder models such as Pegasus and BART, although these models are not performant unless fine-tuned (i.e. many-shot learning). All experiments are done on the test set of datasets (more details in Appendix \ref{appendix:data_model}).

\begin{figure*}[!ht]
  \centering
  \includegraphics[width=0.75\textwidth]{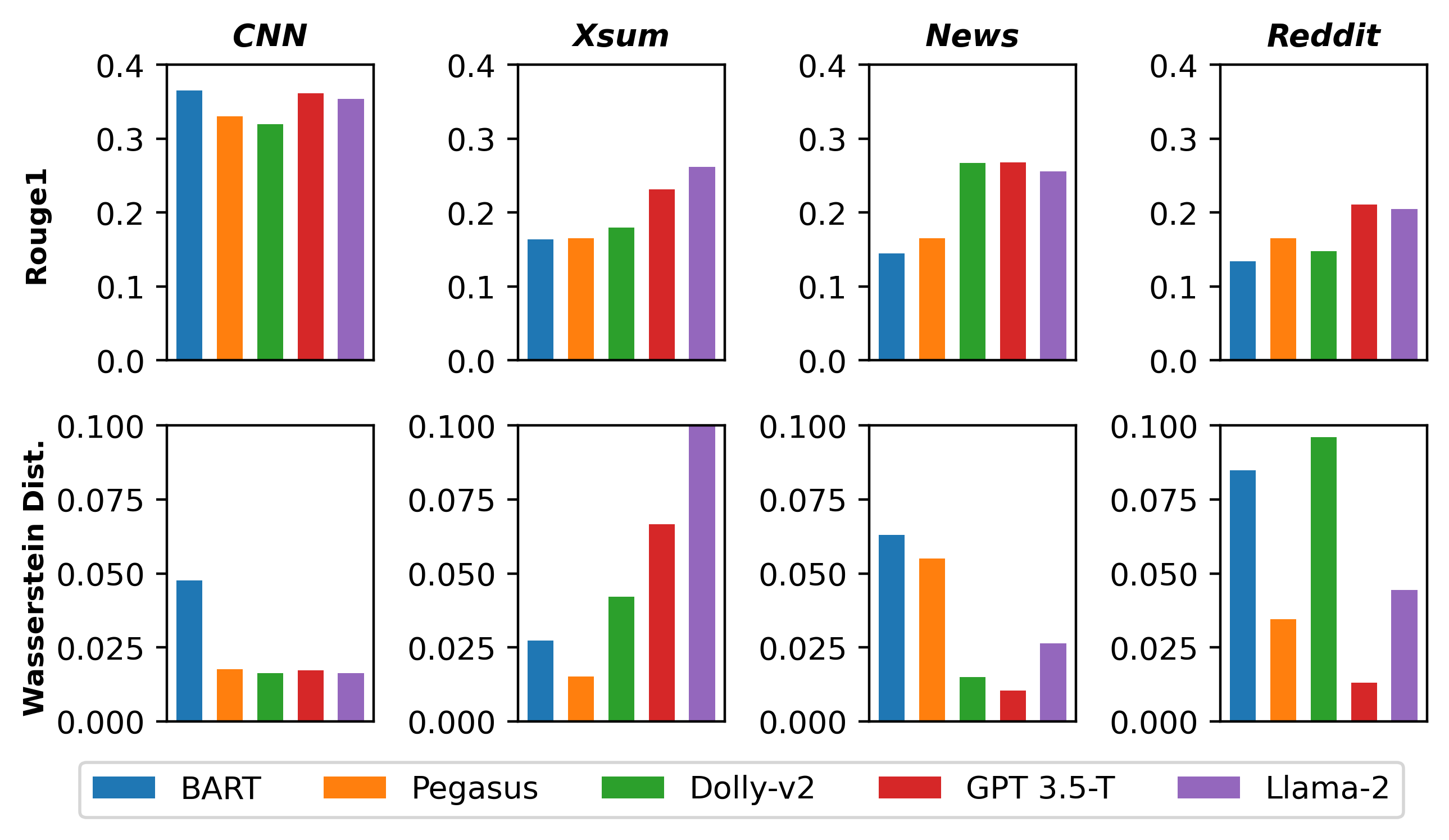}\vspace{-3mm} %0.525
  \caption{Measuring \textit{performance} ($R^1$ score) and \textit{position bias} (Wasserstein distance between gold and generated summaries' positional distributions). Lower Wasserstein distance values correspond to lower position bias.}
  \label{fig:bars}
\end{figure*}

We first visualize the positional distributions generated using our mapping procedure for the gold summaries and model generated summaries in Figure \ref{fig:lines} with $K=10$. As can be seen, \textit{CNN} and \textit{News} contain lead biased gold summaries and \textit{Reddit} and \textit{XSum} are positionally uniformly distributed. It can also be seen that the LLMs tend to have low position bias on \textit{CNN, News,} and \textit{Reddit} datasets. However, for \textit{XSum}, which constitutes the \textit{extreme summarization} setting, LLMs as well as BART/Pegasus exhibit much more lead bias, which is absent from the gold summaries. In \textit{XSum}, the articles are up to 286 sentences long, and summary lengths are required to be between 1-2 sentences long. This is the largest jump from article $\rightarrow$ summary and might explain models' tendency to pick a single sentence from the leading segments of the article. Comparing even with \textit{Reddit} where article lengths are up to 23 sentences long and summaries are 1-17 sentences long, it is evident that \textit{XSum} poses a unique challenge for summarization models.

Next, in Figure \ref{fig:bars}, we measure model performance using the $R^1$ score between gold and generated summaries (results for $R^2, R^L$ are provided in Appendix \ref{appendix:rouge} and follow similar trends) and position bias using Wasserstein distance between the positional distributions of gold and generated summaries of Figure \ref{fig:lines}. As is evident, LLMs attain the highest ROUGE scores on all datasets, and tend to have very low position bias, with the exception of \textit{XSum}. However, even for \textit{XSum}, LLMs achieve excellent performance on ROUGE. BART/Pegasus tend to have low performance and BART is also heavily position biased across all datasets.

\section{Discussion}\label{sec:discussion}

\textbf{Insights on Model Performance and Biases.} %Based on our findings, we observe that:
\begin{itemize}[wide,nosep]
    \item \textit{GPT 3.5-T consistently attains low position bias and high performance.} Generally, \mbox{GPT 3.5-T} should be the de-facto choice for users, as can be seen in Figure \ref{fig:bars}. It consistently obtains high ROUGE scores and low position bias values. However, the paid API and closed-source access might be unfavorable to some users. For open-source models, Llama-2 is the better choice compared to Dolly-v2, and at times obtains ROUGE scores higher than even GPT 3.5-T (for e.g. on \textit{XSum}). In comparison, Dolly-v2 at times has arbitrary and unpredictable performance, such as its low ROUGE scores and large position bias on the \textit{Reddit} dataset, unlike the other LLMs on the same dataset.
    \looseness-1\item \textit{LLMs might exhibit significant lead bias for extreme summarization.} LLMs exhibit strong lead bias in the extreme zero-shot summarization case (Figure \ref{fig:bars}, \textit{XSum}). For users who wish to undertake a similar task (pick 1-2 sentence summaries from very lengthy articles), LLMs might tend to only select sentences/information from the beginning of the article. If this is undesirable, it would be recommended to instead collect gold summaries and finetune LLMs/models to counteract this. 
    \item \textit{Suitability of encoder-decoder models.} As zero-shot summarizers, pretrained encoder-decoder models like BART and Pegasus have high position bias and low performance. This likely stems from their need to be finetuned on article-gold summary tuples to achieve SOTA performance. However, we would like to caution users to ensure that there is no positional mismatch between the data they finetune on and their evaluation set.\footnote{For e.g. finetuning on large \# of news articles collected over the internet to then summarize in a different domain.} While obvious, not ensuring this can lead to low ROUGE scores and high position bias (we demonstrate this in Appendix \ref{appendix:enc_dec}).
\end{itemize}

%\subsection{LLMs and Extreme Summarization}

%\textbf{Suitability of Encoder-Decoder Models.}

\textbf{Choice of Mapping Function $\phi$.} For experiments in the paper we use TF-IDF vector similarities as $\phi$. In our preliminary experiments, we did not observe significant differences for other choices. We provide these additional results when the $R^1$ score is used instead and compare with the original results (Appendix \ref{appendix:phi}). Future work can analyze this choice of $\phi$ as well as effect of other values of $K$.

\looseness-1\textbf{Correlation of Position Bias and ROUGE.} An interesting consequence of our ROUGE and position bias results on datasets shows that their correlation is highly data dependent. For e.g., for \textit{XSum} Spearman's correlation shows statistically significant high positive correlation ($\approx 0.89$) between Wasserstein distances and $R^2$ scores across all models but for \textit{Reddit} there is significant negative correlation ($\approx -0.89$). See Appendix \ref{appendix:correlation} for detailed results. Hence, ROUGE scores are not enough to assess position bias. This also makes intuitive sense as ROUGE simply measures n-gram overlap and cannot holistically evaluate models \cite{cohan2016revisiting}. In future work other evaluation metrics can be studied alongside position bias.

\section{Conclusion}

We analyze zero-shot abstractive summarization by LLMs via a novel formulation of position bias. Position bias measures the tendency of models to generate summaries which overtly and unfairly utilize certain portions of input text over others. Through extensive experiments on the \textit{CNN/DM, XSum, Reddit, News} datasets, as well as various models (GPT 3.5-T, Llama-2, Dolly-v2, Pegasus, BART), we obtain novel insights about model performance and position bias that contribute to a deeper understanding of the challenges and opportunities in leveraging LLMs for effective abstractive summarization.

%\clearpage
\section*{Limitations}

\looseness-1 Our work formulates the concept of position bias in abstractive summarization and analyzes it in LLMs (and other reference models) across four diverse datasets: \textit{CNN/DM, XSum, Reddit,} and \textit{News Summary}. The main limitation is that position bias of LLMs needs to be evaluated on many more datasets, and on other diverse problem settings beyond the ones considered in our paper. Moreover, the source domain itself could be challenging (legal or medical documents) or the LLM might not have been trained with data from that domain. In such cases, the LLM might default to using certain sections of the input articles over others, resulting in position bias. Another limitation of our work has been the primary use of English language datasets, but it is important to benchmark LLM position bias using summarization datasets from other languages as well. Additionally, a limitation of studies on LLMs such as GPT 3.5-Turbo is that they are constantly being updated and improved, and some behaviors might change or become non-existent in future versions \citep{chen2023chatgpt, leiter2024chatgpt}. This necessitates assessing model performance/biases over time. Finally, as preceding ML/AI models are usually designed to be task/domain-specific (e.g. for clustering), issues of bias, fairness, and robustness \citep{iclr2024, iclr2023, neurips2022, afcr2022, ieee2021} specific to these tasks have been naturally studied in the literature. In the same manner, despite their general nature, task-specific robustness/bias needs to be further explored for LLMs.\footnote{Note that independent to our work, \citep{ravaut2023position} obtain similar results as ours for position bias in LLM based summarization.} In future work, we seek to alleviate these limitations.

\section*{Ethics Statement}

Our work on position biases in LLMs is important for understanding how these models prioritize information, and whether or not they disproportionately emphasize specific sections of the source text when generating abstractive summaries in a zero-shot setting. As LLMs are further integrated in society and utilized in various application pipelines, it is crucial to understand their behavior in a transparent manner. Through this study, we wish to shed light on this issue and allow practitioners to understand undersirable model behavior with regards to the summarization task better. This work also enables users to understand scenarios in which these models will generate more reliable outputs, leading to safer outcomes in practice.

\section*{Acknowledgments}
This research was supported in part by NSF Grant ITE 2333736.

\bibliography{refs}
\bibliographystyle{acl_natbib}

%\clearpage

\appendix

\section*{Appendix}

\section{Dividing Articles into $K$ Segments of (Approximately) Equal Length} \label{appendix:math}
To overcome the issue of articles being of differing lengths, we need better mathematical structure for describing \textit{position} across articles in a dataset. For this, we wish to divide an article $x$ into $K$ segments of approximately equal length. To achieve this, the $j$-th segment will contain the sentences of the article that lie in the interval:\\ $[(j-1)\cdot c + \min(j-1,d), j\cdot c + \min(j,d)-1]$, \\where $K \leq \min_{x \in X}|x|$, $c = \floor*{\frac{N_x}{K}}$, and ${d = N_x \mod K}$. 

\looseness-1The aim is to distribute the content of the article into $K$ segments in a way that makes the lengths of these segments as equal as possible. Here, the inequality $K \leq \min_{x \in X}|x|$ ensures that the number of desired segments $K$ should not exceed the length of the shortest article in the set of articles $X$ (otherwise it will lead to empty segments for those articles). The content of each segment $j$ (note, $j$ represents the index of the segment from 1 to $K$) is determined by an interval defined by: $(j-1)\cdot c + \min(j-1,d)$ (lower bound) and $j\cdot c + \min(j,d)-1$ (upper bound). 

Intuitively, $c = \floor*{\frac{N_x}{K}}$ calculates the approximate length of each segment as it divides the total number of sentences in the article ($N_x$) by the desired number of segments ($K$) and rounds down to the nearest whole number. However, $N_x$ might not be fully divisible by $K$ and hence, we might have remainder $d = N_x\mod K$. Hence, $d$ accounts for any additional content that cannot be evenly distributed among the segments and ensures that segments accommodate the variation in article lengths. In this manner, the terms $\min(j-1,d)$ and $\min(j,d)$ in the lower and upper bounds of the interval are used to account for potential variations in segment length due to the remainder $d$.

\section{Additional Results for Other ROUGE Metrics}\label{appendix:rouge}

\begin{figure}[!ht]
  \centering
  \includegraphics[width=0.48\textwidth]{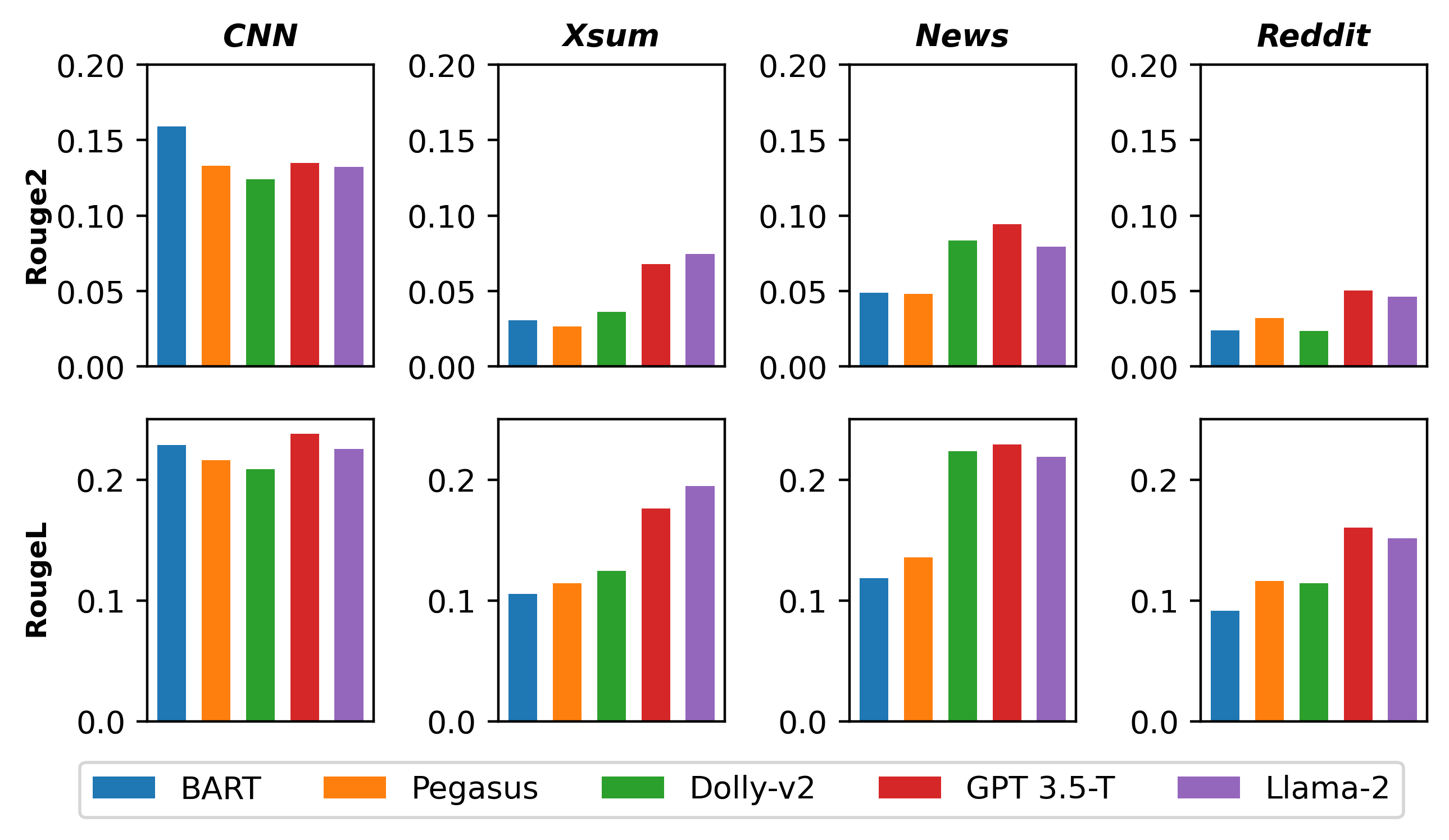}\vspace{-3mm} %0.525
  \caption{Additional results for $R^2$ and $R^L$ metrics.}
  \label{fig:appendix_bars}
\end{figure}

In the main paper in Figure \ref{fig:bars} we provided results for the ROUGE-1 ($R^1$) score. Here, we provide additional results for the ROUGE-2 ($R^2$) and ROUGE-L ($R^L$) scores measured between the gold and model generated summaries as Figure \ref{fig:appendix_bars}. It can be seen that the trends are similar to $R^1$ and LLMs exhibit stellar performance for $R^2$ and $R^L$ across all datasets.

\section{Additional Position Bias Results for Finetuning BART and Pegasus}\label{appendix:enc_dec}

\begin{figure*}[!ht]
  \centering
  \includegraphics[width=0.625\textwidth]{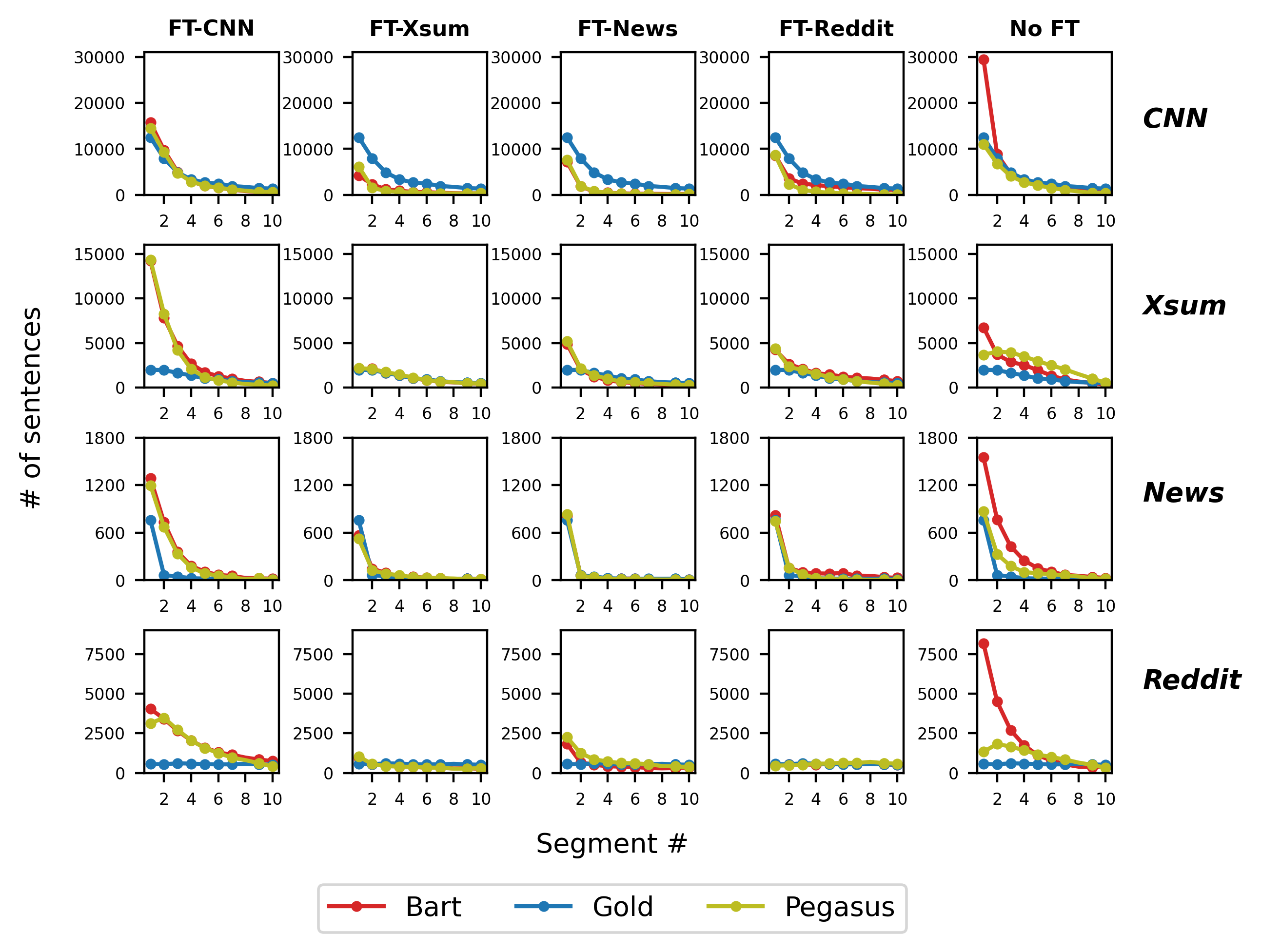}\vspace{-4mm}
  \caption{Visualizing positional distributions of gold and Pegasus/BART generated summaries for all datasets with and without finetuning on a particular dataset (training set). For the finetuned models, the diagonal subfigures are the ones that have the same finetuning and evaluation datasets and have low position bias. All other subfigures have a \textit{mismatch} between finetuning and evaluation datasets, and exhibit high levels of position biases. That is, the model generated summary positional distribution is very different from the gold summary positional distribution. The no-finetuning results were also shown in Figure \ref{fig:lines} and are provided again for reference.}
  \label{fig:appendix_lines}
\end{figure*}

We go beyond the zero-shot setting to provide additional results on measuring position bias when BART and Pegasus are finetuned on the datasets we consider. The training was carried out on one \textit{NVIDIA-A100} with 50 GB memory. We use the HuggingFace Seq2Seq Trainer Class with a batch size of 64, gradient checkpointing of 4 and gradient accumulation. We use mixed-precision training for all models. The learning rate for all models was set to \textit{5.6e-5}. While generating summaries during finetuning we use a single beam and maximum generation length of 128. 

We finetune on the training set of each of the 4 datasets and evaluate on all datasets (for reference we again provide the no-finetuning / zero-shot results of Figure \ref{fig:lines}). Results for the obtained positional distributions are shown in Figure \ref{fig:appendix_lines}. It is evident that if there is mismatch in the finetuning and evaluation datasets for pretrained encoder-decoder models, they exhibit high position bias, leading to biased summarization. Hence, it is important for practitioners to collect article-summary data for finetuning that exactly reflects their evaluation or production use-case.

\section{Additional Results for Different $\phi$}\label{appendix:phi}

\begin{figure}[!ht]
  \centering
  \includegraphics[width=0.48\textwidth]{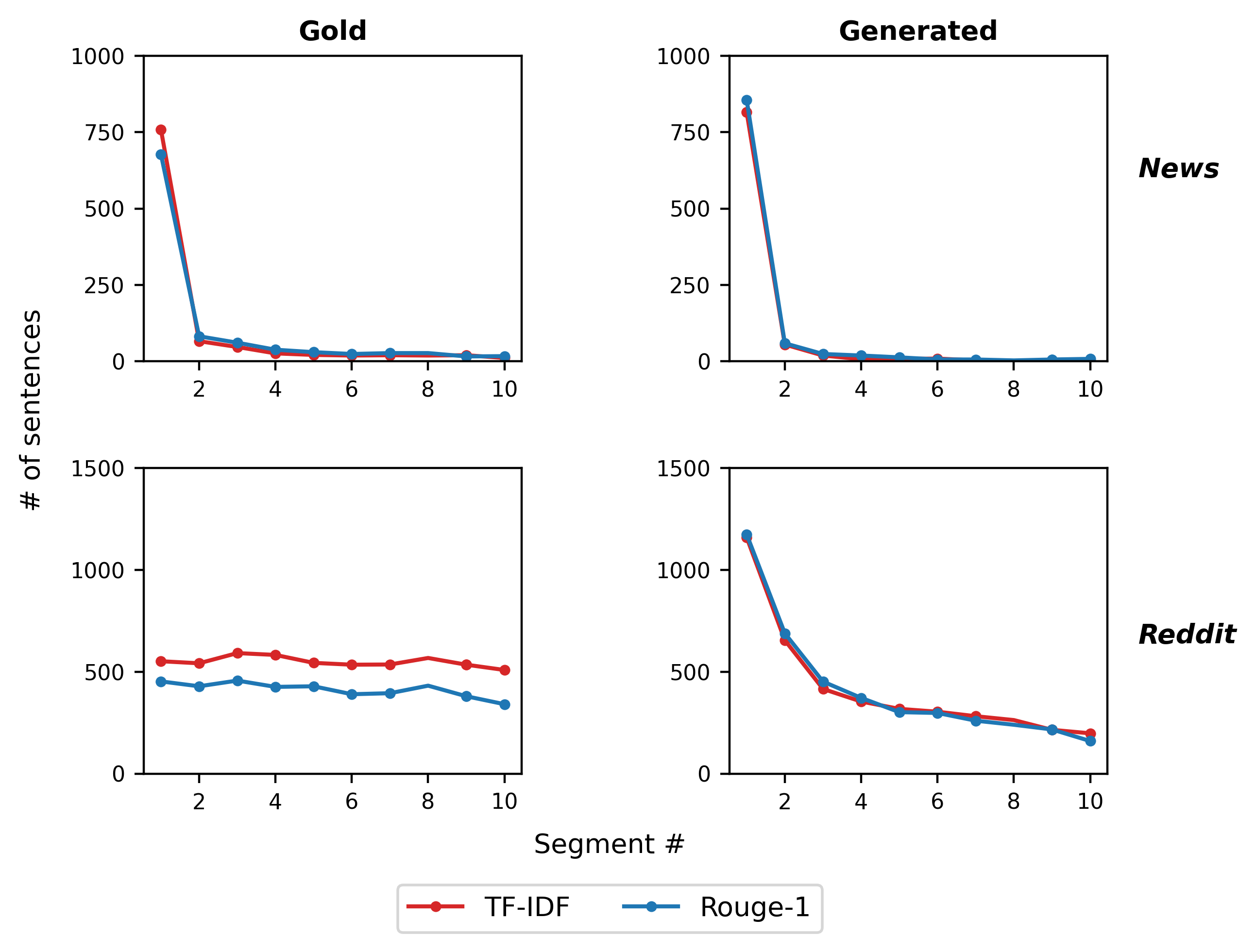}\vspace{-3mm} %0.525
  \caption{Results on \textit{News} and \textit{Reddit} for Llama-2 when $\phi$ is either TF-IDF similarity or ROUGE-1.}
  \label{fig:appendix_rouge}
\end{figure}

\looseness-1For experiments in the main paper, we opt for TF-IDF vector similarities as the choice of the mapping function $\phi$ due to computational efficiency (over computing individual ROUGE scores between summary and article sentences for e.g.). However, it is important to examine whether this choice significantly impacts results, trends, and our findings. In initial experiments with different $\phi$ we concluded that this choice does not affect results. In Figure \ref{fig:appendix_rouge} we provide results that support this by using $R^1$ (ROUGE-1) as the metric for $\phi$ on the \textit{Reddit} and \textit{News} datasets for Llama-2 generated summaries. We compare the gold summary and generated summary positional distributions for both datasets when $\phi$ is computed using TF-IDF vectors and $R^1$. It is clear that the trends and results are the same for both $\phi$. Even the Wasserstein distance values computed between gold and generated summaries do not change much. For e.g. on \textit{Reddit}: for TD-IDF the distance value is 0.044 and for $R^1$ it is 0.046. Despite no significant differences, we believe future work can explore the choice of $\phi$ more deeply.

\section{Additional Results for Measuring Correlation Between ROUGE and Position Bias}\label{appendix:correlation}

\begin{table}[!ht]
\centering
\caption{Measuring Spearman's correlation coefficient between position bias (Wasserstein distances) and ROUGE metrics for all datasets (* denotes p-values of $\leq0.1$ and ** denotes p-values of $\leq 0.05$).}\label{table:corr}
\resizebox{0.3\textwidth}{!}{%
\begin{tabular}{ccc} 
\hline
\textbf{Dataset} & \textbf{Metric} & \textbf{Correlation} \\  
\hline
\multirow{3}{*}{\textit{CNN/DM}} & $R^1$ & 0.499 \\
& $R^2$ & 0.799\textsuperscript{*} \\
& $R^L$ & 0.300 \\
\hline
\multirow{3}{*}{\textit{XSum}} & $R^1$ & 0.899\textsuperscript{**} \\
& $R^2$ & 0.999\textsuperscript{**} \\
& $R^L$ & 0.899\textsuperscript{**} \\
\hline
\multirow{3}{*}{\textit{News}} & $R^1$ & -0.999\textsuperscript{**} \\
& $R^2$ & -0.899\textsuperscript{**} \\
& $R^L$ & -0.999\textsuperscript{**} \\
\hline
\multirow{3}{*}{\textit{Reddit}} & $R^1$ & -0.799\textsuperscript{*} \\
& $R^2$ & -0.899\textsuperscript{**} \\
& $R^L$ & -0.799\textsuperscript{*} \\
\hline
\end{tabular}}
\end{table}

In this section, we measure the correlation between ROUGE scores \cite{lin2004rouge} and Wasserstein distance computed between the gold summary and model generated summary distributions. We conduct this experiment using Spearman's correlation coefficient statistic over all models and for each dataset. We utilize the $R^1, R^2, R^L$ ROUGE metrics individually for this analysis, and the results are shown in Table \ref{table:corr}. We find that correlation is highly dependent on the dataset: for \textit{CNN} the correlation is not strong and the results are not statistically significant, for \textit{XSum} ROUGE and position bias are positively correlated and statistically significant, and for \textit{News} and \textit{Reddit} results are statistically significant but highly negatively correlated. This indicates that ROUGE itself is not enough to assess position bias and hence, independent position bias measurement is important for holistic summarization evaluation.

\section{Dataset, Model, and Training Details}\label{appendix:data_model}

\subsection{Detailed Dataset Information} %give all details of datasets, num sentences in articles summaries, avg sentences, where dataset comes from etc etc.
\textbf{\textit{XSum}} \citep{xsum}: The \textit{XSum} dataset contains over 200K short, one-sentence news summaries answering the question "What is the article about?" and was collected by harvesting online articles from the British Broadcasting Corporation (BBC). The testing set consists of 11334 articles. The average number of sentences in the articles are 19.105. The total number of sentences in the summaries are 11334, leading to an average of 1 sentence per summary.

\noindent\textbf{\textit{CNN/DM}} \citep{cnn}:  The \textit{CNN/DM} dataset contains 300K unique news articles as written by journalists at CNN and the Daily Mail and is one of the most popular datasets for abstractive/extractive summarization and abstractive question answering. The testing set consists of 11490 articles. The average number of sentences in the articles was 33.37. The total number of sentences in the summaries was 43560 (an average of 3.79 sentences per summary).

\noindent\textbf{\textit{Reddit TL;DR}} \citep{reddit}: The \textit{Reddit} dataset consists of 120K posts from the online discussion forum Reddit. The authors used these informal crowd-generated posts as text source, in contrast with existing datasets that mostly use formal documents as source such as news articles. We used an 80-20\% train-test split to obtain 4214 articles in the test set. The average number of sentences per article was 22.019. The total number of sentences in the summaries was 6016 which leads to an average of 1.4276 sentences per summary.

\noindent\textbf{\textit{News Summary}} \citep{news}: The \textit{News} dataset was initially created for fake news classification. We used the testing set comprising of 1000 articles. The number of sentences in the summaries are 1012 (an average of 1.012 per summary)

\subsection{Models}
\noindent\textbf{\textit{Pegasus}} \citep{zhang2020pegasus}: The Pegasus model family is used mainly for text-summarization tasks. We use the \textit{google/pegasus-large} checkpoint\footnote{\scriptsize\url{https://huggingface.co/google/pegasus-large}} from Huggingface as the summarization model.

\noindent\textbf{\textit{BART}} \citep{lewis2020bart}: BART is a Seq2Seq encoder-decoder model for language tasks. We use the \textit{facebook/bart-large} checkpoint\footnote{\scriptsize\url{https://huggingface.co/facebook/bart-large}} from Huggingface as the summarization model.

\noindent\textbf{\textit{GPT 3.5-T}} \footnote{\scriptsize\url{https://platform.openai.com/docs/models/gpt-3-5}}:
GPT-3.5-turbo is OpenAI's flagship LLM which has been instruction-tuned and optimized for chat purposes. We utilized the model from Microsoft Azure's OpenAI service and the version was the August 3rd version.

\noindent\textbf{\textit{Llama2-13B-chat}} \footnote{\scriptsize\url{https://huggingface.co/meta-llama/Llama-2-13b-chat-hf}}: Meta developed and publicly released the Llama-2 family of LLMs, a collection of pretrained and fine-tuned generative text models ranging in scale from 7-70B parameters. The chat versions of the models are optimized for dialogue via instruction finetuning. We generated inferences by modifying the PyTorch code provided in the official Github repository: \url{https://github.com/facebookresearch/llama}.

\noindent\textbf{\textit{Dolly-v2-7B}} \citep{dolly}: Dolly-v2-7B is a 6.9 billion parameter causal language model created by Databricks that is derived from EleutherAI's Pythia-6.9B model and finetuned on a 15K instruction corpus generated by Databricks employees. We used the \textit{databricks/dolly-v2-7b} checkpoint\footnote{\scriptsize\url{https://huggingface.co/databricks/dolly-v2-7b}} from HuggingFace.

\subsection{Generating Summaries via LLMs}

\looseness-1We provide the prompts used to generate summaries for each LLM and each dataset (prompts might differ between datasets for the same model due to different summary requirements, and they might differ across models as different models respond to input text differently). Note that \textit{\{Article\}} in each prompt should be replaced by the article to be summarized. It is also important to note that the prompts were adapted iteratively through multiple experiments to ensure that models followed the prompt as closely as possible. At times models did not follow the prompt specifications exactly and would generate more summary sentences than required for that dataset (for e.g. GPT 3.5-T followed exact prompt specifications 74.9\% of the time). Hence, for parity between dataset and model summaries, and fair comparison between all models, we uniformly randomly sampled (so as to not add inductive bias) the number of sentences required from the generated output. Also, due to OpenAI's content moderation policy GPT 3.5-T did not give responses for a minority of inputs (6.16\% of all input). We believe future LLM versions will improve along these lines to always follow prompts exactly as specified so post-hoc measures will not be required. We now provide prompts below.

\subsubsection{Prompts for GPT 3.5-T}\label{appendix:gpt_prompts}
\noindent\textbf{\textit{Xsum}}: \small\textit{For the following article: \{Article\}. Return a summary comprising of 1 sentence. Write the sentence in a dash bulleted format.}\normalsize

\noindent\textbf{\textit{CNN/DM}}: \small\textit{For the following article: \{Article\}. Return a summary comprising of 3 sentences. Write each sentence in a dash bulleted format.}\normalsize

\noindent\textbf{\textit{Reddit}}: \small\textit{For the following article: \{Article\}. Return a summary comprising of 1 sentence. Write the sentence in a dash bulleted format.}\normalsize

\noindent\textbf{\textit{News}}: \small\textit{For the following article: \{Article\}. Return a summary comprising of 1 sentence. Write the sentence in a dash bulleted format.}\normalsize

\subsubsection{Prompts for Llama2-13B-chat}\label{appendix:llama_prompts}
\noindent\textbf{\textit{Xsum}}: \small\textit{For the following article: \{Article\}. Return a summary comprising of 1 sentence. Write the sentence in a numbered list format.\\For example:\\1. First sentence}\normalsize

\noindent\textbf{\textit{CNN/DM}}: \small\textit{For the following article: \{Article\}. Return a summary comprising of 3 sentence. Write the sentence in a numbered list format.\\For example:\\1. First sentence\\2. Second sentence\\3. Third sentence}\normalsize

\noindent\textbf{\textit{Reddit}}: \small\textit{For the following article: \{Article\}. Return a summary comprising of 1 sentence. Write the sentence in a numbered list format.\\For example:\\1. First sentence}\normalsize

\noindent\textbf{\textit{News}}: \small\textit{For the following article: \{Article\}. Return a summary comprising of 1 sentence. Write the sentence in a numbered list format.\\For example:\\1. First sentence}\normalsize

\subsubsection{Prompts for Dolly-v2-7B}\label{appendix:dolly_prompts}
\noindent\textbf{\textit{Xsum}}: \small\textit{Generate a 1 sentence summary for the given article. Article: {\{Article\}}.}\normalsize

\noindent\textbf{\textit{CNN/DM}}: \small\textit{Generate a 3 sentence summary for the given article. Article: {\{Article\}}. }\normalsize

\noindent\textbf{\textit{Reddit}}: \small\textit{Generate a 1 sentence summary for the given article. Article: {\{Article\}}. }\normalsize

\noindent\textbf{\textit{News}}: \small\textit{Generate a 1 sentence summary for the given article. Article: {\{Article\}}. }\normalsize

% \subsection{General Remarks \td} %put info about capping strategy etc here, anything that is relevant to LLM output

% The prompts were adapted based off what was returning the highest rate of prompt following. Due to OpenAI's content moderation policy GPT 3.5-T gave responses for 93.84\% of the prompts. The rest were met with the message "The response was filtered due to the prompt triggering Azure OpenAI's content management policy. Please modify your prompt and retry." 

% Additionally, GPT 3.5-T followed the instructions in the prompt i.e outputting the exact number of summary sentences 74.9\% of the time. Llama2-13B-chat followed the instructions 45.99\% of the instances and Dolly-v2-7b followed the instructions a in a mere 20.77\% of the cases. 

% Since most of the time the models output more summary sentences than asked of them, we decided to truncate the summaries to follow the original prompt. This would lead to a fair comparison since there would be an equivalent number of sentences in the Gold and Generated summaries. We performed this truncation in the Xsum and News datasets for the GPT 3.5-T model. For the Xsum, News and Reddit datasets in Llama-13b-chat and Dolly-v2-7b models. We decided to randomly pick a sentence from the generated sentences as it would prevent adding any unforeseen biases to the generated summaries. 

\section{Analyzing the Effect of Prompt Engineering Methods}
\begin{figure}[!ht]
  \centering
  \includegraphics[width=0.42\textwidth]{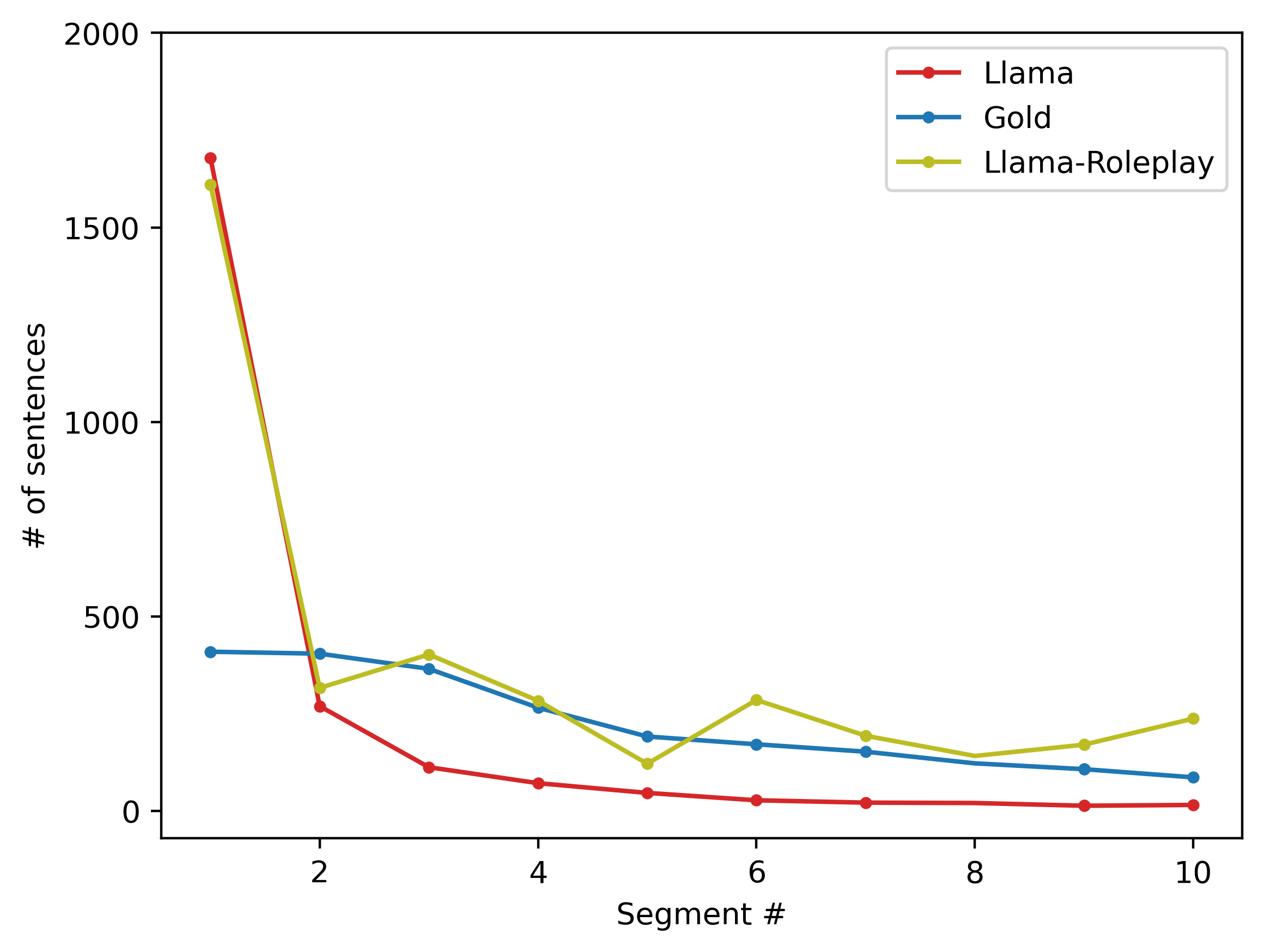}\vspace{-3mm} %0.525
  \caption{Using role-playing on Llama-2 and \textit{XSum}.}
  \label{fig:appendix_roleplay}
\end{figure}

To motivate future work and showcase the generalizability of our framework, we include results for position bias when a zero-shot prompt engineering approach is used: \textit{role-playing} \citep{roleplay}. Role-playing has been shown to effectively increase LLM performance and reasoning abilities. We randomly sampled 2275 articles from the \textit{XSum} dataset and utilize the \textit{Llama2-13B-chat} LLM. Then, we use each of the 2275 articles to plot position bias distributions for summaries generated using role-playing and our original prompt generated summaries (as well as gold summaries) for comparison. This result is shown as Figure \ref{fig:appendix_roleplay}.

It can be seen that the curves change slightly-- and the role-play summary distribution becomes closer to the gold summary distribution, as expected. However, the overall trends are similar as lead bias is still prominent. Clearly, role-playing on this subset of data is not an exhaustive study, but future work can expand on uncovering how prompt engineering methods (e.g. role-playing, among others) affect summarization position bias.

\section{Additional Results on Flan-T5}
\begin{figure}[!ht]
  \centering
  \includegraphics[width=0.33\textwidth]{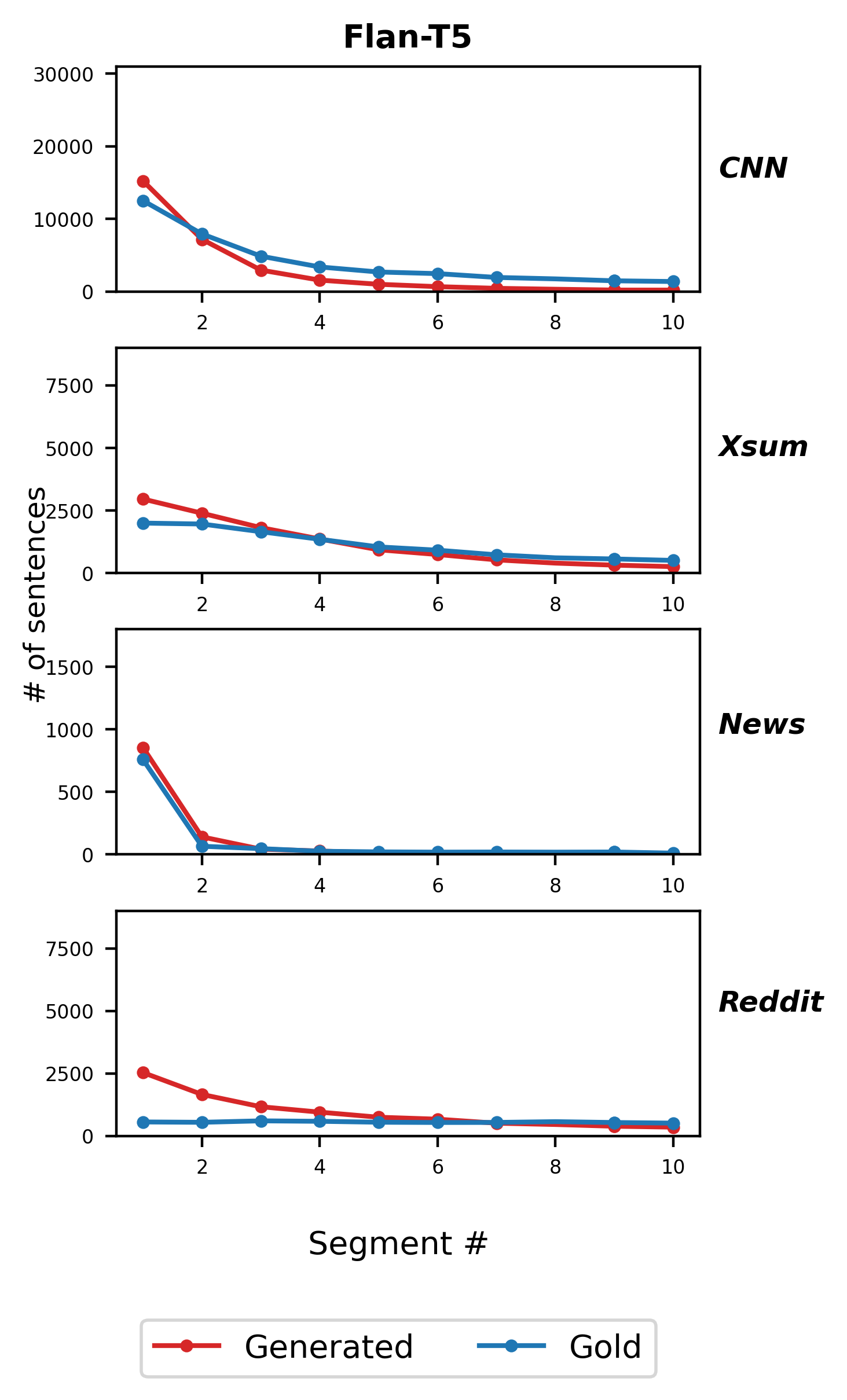}\vspace{-3mm} %0.525
  \caption{Position bias results for Flan-T5.}
  \label{fig:appendix_flan}
\end{figure}

Since we have primarily considered specialized encoder-decoder models such as BART and Pegasus in this work, we also provide additional results for position bias when a generalized encoder-decoder model such as Flan-T5 \citep{flant5} is used instead. These results can be observed in Figure \ref{fig:appendix_flan} for all 4 of our datasets. As can be seen in the figure, position bias is low for all datasets, and especially \textit{XSum} (which contrasts with LLMs). This is also observable in the Wasserstein distance values ($\approx$0.050 for \textit{Reddit} and \textit{CNN/DM}, 0.024 for \textit{News}, and only 0.015 for \textit{XSum}).

\section{Mapping Summary Sentences to Multiple Article Sentences}
Currently, $\phi$ maps back from one summary sentence to one article sentence that contributes the most to that summary sentence. To do this, as $\phi$ measures similarity between sentences, we currently only pick the article sentence with the maximum similarity to the summary sentence. However, since $\phi$ is basically measuring similarity, we can return the top-2 or top-3 matches and undertake the same position bias analysis as in the main results. No specific change is necessary, since our position bias estimation is done in aggregate, via binning. It is beneficial to assess the impact of utilizing multiple article sentences, especially for datasets like \textit{XSum} where the summary is usually just one sentence and discusses facts from multiple article sentences.

\begin{figure}[!ht]
  \centering
  \includegraphics[width=0.49\textwidth]{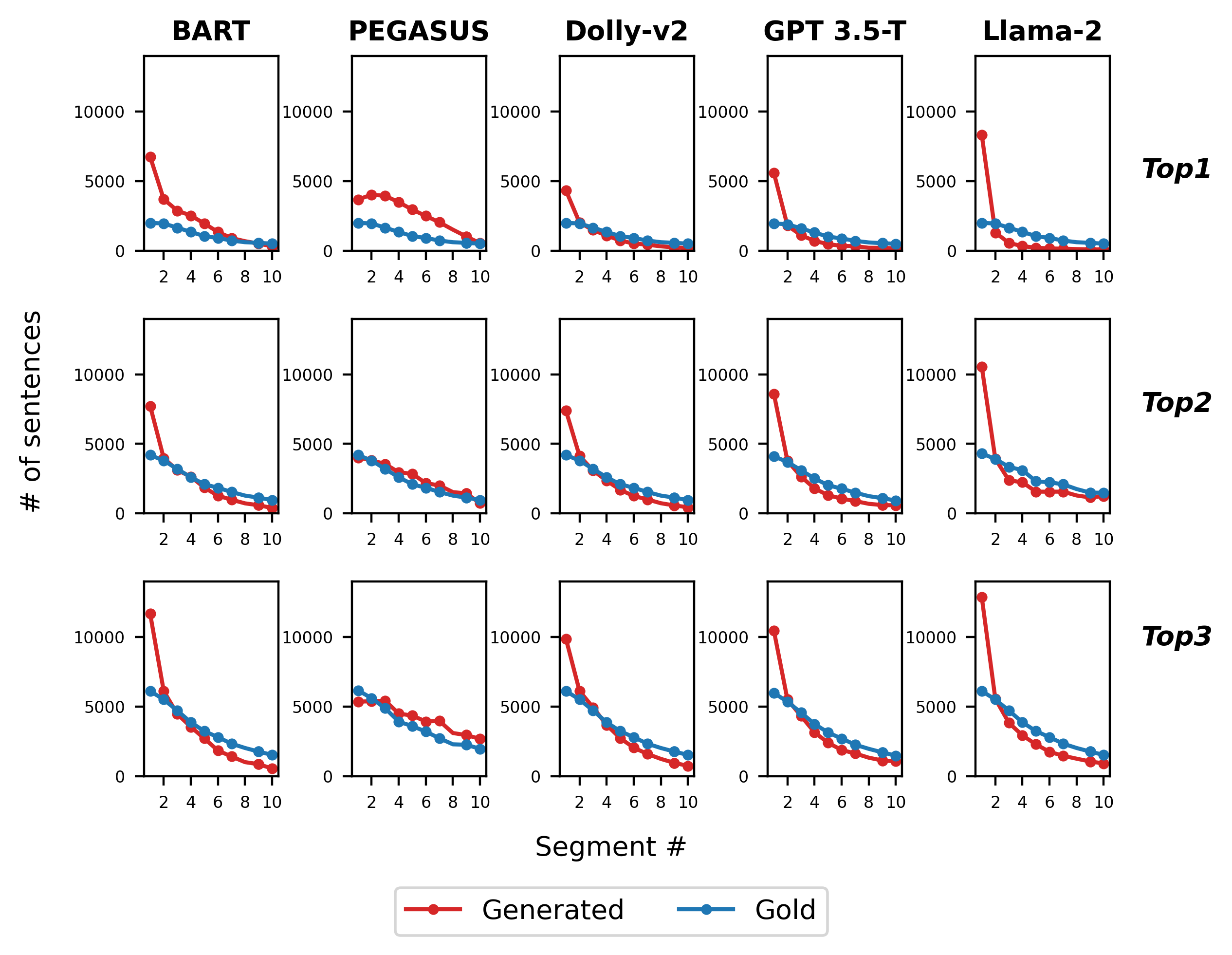}\vspace{-3mm} %0.525
  \caption{Mapping summary sentences to multiple article sentences for measuring position bias on \textit{XSum}.}
  \label{fig:appendix_topn}
\end{figure}

We undertake this analysis for each of our 5 models in the paper on the \textit{XSum} dataset and the results are shown in Figure \ref{fig:appendix_topn}. Here, we have provided position bias distributional results for only using the top-1 match (our original results), top-2 matches, and top-3 matches. It can be seen that the distributions do change slightly, but overall the trends remain the same. More specifically, lead bias for each of the LLMs on \textit{XSum} is further exacerbated, indicating that even the top-1 match provides good estimates for position bias.

\section{Code and Reproducibility}\label{appendix:code}

We open-source our code and provide it as a Github repository: \url{https://github.com/anshuman23/LLM_Position_Bias}. The repository contains explicit instructions for how to reproduce our results and analyze the findings for each model. We used Python 3.8.10 for all experiments. The experiments were conducted on Ubuntu 20.04 using NVIDIA GeForce RTX A6000 GPUs running with CUDA version 12.0.

\end{document}